\documentclass[conference]{IEEEtran}
\IEEEoverridecommandlockouts

\usepackage{cite}
\usepackage{amsmath,amssymb,amsfonts}
\usepackage{graphicx}
\usepackage{textcomp}
\usepackage{xcolor}
\usepackage{hyperref}
\usepackage{leftindex}

\usepackage{algorithm}
\usepackage{algpseudocode}
\usepackage{subcaption}

\usepackage{balance}

\DeclareMathOperator*{\argmax}{argmax}

\def\BibTeX{{\rm B\kern-.05em{\sc i\kern-.025em b}\kern-.08em
    T\kern-.1667em\lower.7ex\hbox{E}\kern-.125emX}}

\makeatletter
\newcommand{\linebreakand}{%
  \end{@IEEEauthorhalign}
  \hfill\mbox{}\par
  \mbox{}\hfill\begin{@IEEEauthorhalign}
}
\makeatother 
    
\begin{document}

\title{Active Robot Curriculum Learning from Online Human Demonstrations
}

\author{\IEEEauthorblockN{Muhan Hou}
\IEEEauthorblockA{\textit{Vrije Universiteit Amsterdam}\\
\textit{Department of Computer Science}\\
Amsterdam, the Netherlands \\
m.hou@vu.nl}
\and
\IEEEauthorblockN{Koen Hindriks}
\IEEEauthorblockA{\textit{Vrije Universiteit Amsterdam}\\
\textit{Department of Computer Science}\\
Amsterdam, the Netherlands \\
k.v.hindriks@vu.nl}
\linebreakand

\IEEEauthorblockN{A.E. Eiben}
\IEEEauthorblockA{\textit{Vrije Universiteit Amsterdam}\\
\textit{Department of Computer Science}\\
Amsterdam, the Netherlands \\
a.e.eiben@vu.nl}
\and
\IEEEauthorblockN{Kim Baraka}
\IEEEauthorblockA{\textit{Vrije Universiteit Amsterdam}\\
\textit{Department of Computer Science}\\
Amsterdam, the Netherlands \\
k.baraka@vu.nl}
}

\maketitle

\begin{abstract}
Learning from Demonstrations (LfD) allows robots to learn skills from human users, but its effectiveness can suffer due to sub-optimal teaching, especially from untrained demonstrators. Active LfD aims to improve this by letting robots actively request demonstrations to enhance learning. However, this may lead to frequent context switches between various task situations, increasing the human cognitive load and introducing errors to demonstrations. Moreover, few prior studies in active LfD have examined how these active query strategies may impact human teaching in aspects beyond user experience, which can be crucial for developing algorithms that benefit both robot learning and human teaching. To tackle these challenges, we propose an active LfD method that optimizes the query sequence of online human demonstrations via Curriculum Learning (CL), where demonstrators are guided to provide demonstrations in situations of gradually increasing difficulty. We evaluate our method across four simulated robotic tasks with sparse rewards and conduct a user study ($N=26$) to investigate the influence of active LfD methods on human teaching regarding teaching performance, post-guidance teaching adaptivity, and teaching transferability. Our results show that our method significantly improves learning performance compared to three other LfD baselines in terms of the final success rate of the converged policy and sample efficiency. Additionally, results from our user study indicate that our method significantly reduces the time required from human demonstrators and decreases failed demonstration attempts. It also enhances post-guidance human teaching in both seen and unseen scenarios compared to another active LfD baseline, indicating enhanced teaching performance, greater post-guidance teaching adaptivity, and better teaching transferability achieved by our method.
\end{abstract}


\section{Introduction}
Learning from Demonstrations (LfD) is a widely used approach that allows robots to acquire task skills by imitating humans perform these tasks. Traditional LfD methods \cite{ball2023efficient, nair2020awac, nair2018overcoming, vecerik2017leveraging}, however, rely on offline demonstrations collected prior to training. When these demonstrations are selected and provided by everyday human users, the robot learning may suffer from sub-optimal human teaching (e.g., providing demonstrations with a biased distribution) \cite{hou2023shaping}. To alleviate this issue, recent approaches \cite{hou2024give, chen2020active, chenactive} have enabled robots to actively query for online demonstrations during the learning process, leading to improved policy performance and sample efficiency.

\begin{figure}[t!]
  \centering
  \includegraphics[width=1.0\linewidth]{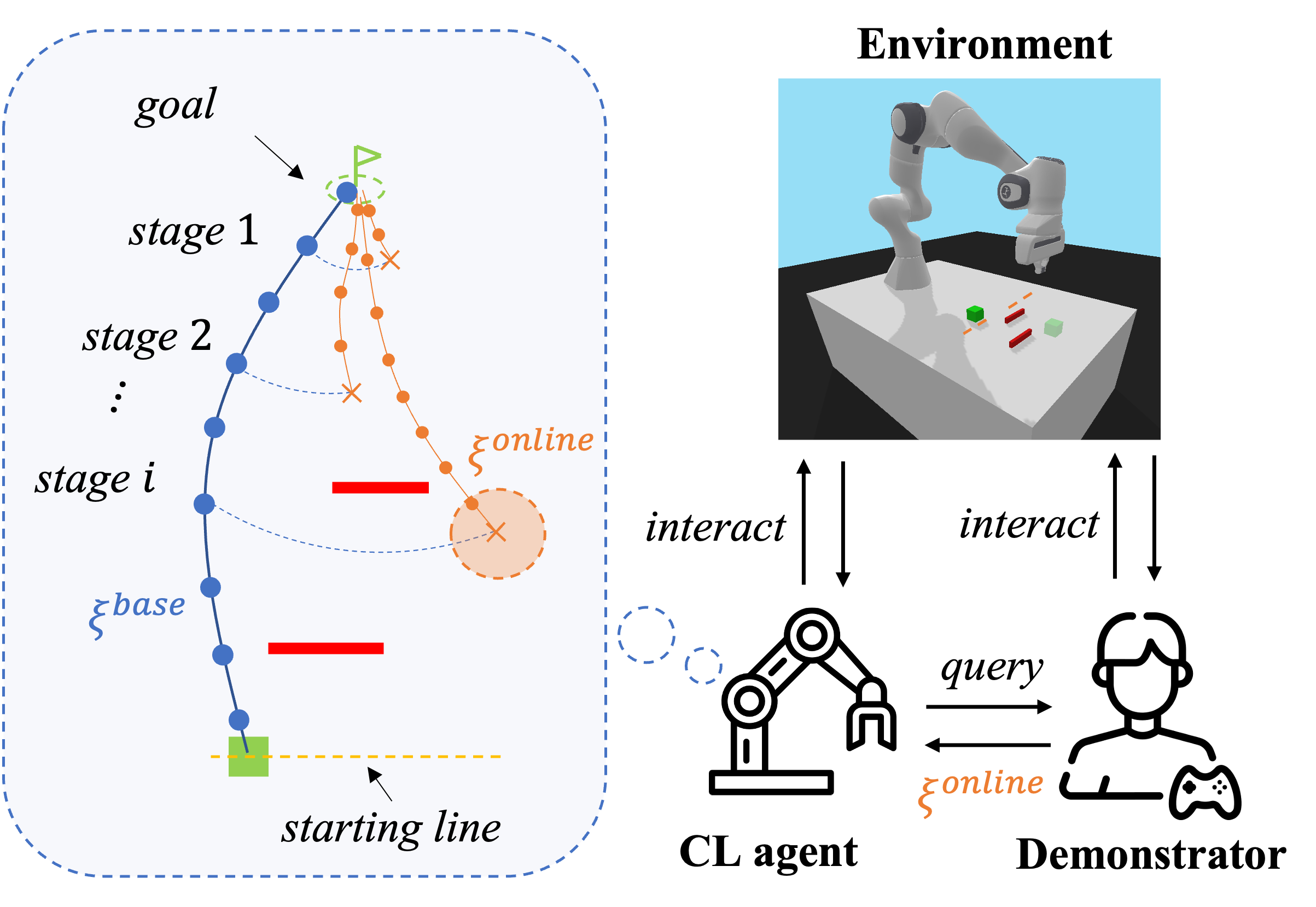}
  \vspace{-6mm}
  \caption{Overview of our method. We present an active LfD algorithm that optimizes the query sequence of online human demonstrations via Curriculum Learning (CL), where human demonstrators are guided to provide online demonstrations in situations of gradually increasing difficulty.}
  \label{overview}
  \vspace{-3mm}
\end{figure}

Although active LfD approaches improve learning performance, they come at the cost of greater human effort. Many of these methods require resetting the environment to specific states whenever the learning agent encounters uncertainty \cite{chen2020active, chenactive, hoque2021lazydagger, kelly2019hg, ross2011reduction}, prompting the human to assist with the next steps. This process results in frequent and abrupt context switches across different areas of the task space, imposing significant cognitive load on the human demonstrator and potentially introducing errors or noise into the demonstrations \cite{hou2024give}. Moreover, few prior studies in active LfD have examined how these active query strategies may impact human teaching beyond user experience. Important factors, such as shifts in teaching strategies after robot guidance and the ability of human teachers to transfer guidance to unseen scenarios, have been underexamined, yet they are crucial for developing algorithms that benefit both robot learning and human teaching.

Aiming to benefit both robot learning and human teaching, we present an active LfD algorithm that leverages online human demonstrations for policy updates. More specifically, our work is inspired by Curriculum Learning (CL) \cite{bengio2009curriculum}, the general idea of which is to break the original hard task into a sequence of sub-tasks with gradually increasing difficulties. CL situates the learning agent in the next sub-task only after it masters the simpler ones, proven to be able to improve learning performance \cite{zhang2020automatic, florensa2017reverse, ivanovic2019barc, beyret2019dot, dai2021automatic, taoreverse}. This scaffolding process is intuitive to common human users as it well aligns with how humans learn to perform a new task \cite{krueger2009flexible, skinner1961teaching}. Furthermore, the gradual progression between related sub-tasks may reduce the human effort required to adapt to context switches, as robot queries are presented in scenarios that are similar to one another but incrementally more challenging.

Building on these insights, we design an active LfD algorithm (Fig. \ref{overview}) that optimizes the sequence of demonstration queries with gradually increasing difficulty. These queries not only gather demonstrations for the robot to imitate but also automate the curriculum expansion and iteration process, guiding policy exploration. We evaluate our method on four simulated robotic tasks with sparse rewards and conduct a user study to investigate how our method may benefit human teaching. Compared to three other LfD baselines, our method outperformed them in all four tasks in terms of converging policy performance and sample efficiency. Regarding human teaching, compared with another active LfD baseline, our method takes significantly less human time with much fewer failed demonstration attempts during demonstration collection, and achieves much better post-guidance human teaching in both seen and unseen tasks. These results indicate the improved per-guidance teaching performance, post-guidance teaching adaptivity, and teaching transferability of human teaching achieved by our method. To summarize, the main contributions of our work are as follows:
\begin{itemize}
    \item An active LfD algorithm that leverages Curriculum Learning to optimize the query sequence of online human demonstrations, leading to significant improvements in learning performance on various sparse-reward robotic tasks. 
    \item A user study that investigates how human teaching, including per-guidance teaching performance, post-guidance teaching adaptivity, and teaching transferability, can be influenced by active LfD algorithms.
\end{itemize}

\section{Related Work}
\subsection{Curriculum Learning with Demonstrations}
Curriculum Learning (CL) \cite{bengio2009curriculum} is a method that breaks down the original task into a sequence of subtasks of increasing difficulty and presents them to the learning agent incrementally once it has mastered the simpler ones. This scaffolding process has been shown to improve learning performance \cite{bengio2009curriculum}, especially for robotic tasks with sparse rewards \cite{zhang2020automatic, florensa2017reverse, ivanovic2019barc, beyret2019dot}. To further automate curriculum design, efforts have been made to utilize offline demonstrations to guide curriculum iteration. For instance, \cite{dai2021automatic} leverages offline expert demonstrations to divide the task space into discrete stages, iteratively resetting the environment to the area near the goal space and progressively exploring backward to the initial state space. Similarly, \cite{hermann2020adaptive} employs a reverse curriculum strategy and enhances the curriculum update rules by incorporating simulation environment parameters to improve adaptability for sim-to-real applications. Additionally, \cite{taoreverse} combines the reverse curriculum with the forward curriculum after the agent masters the task for all the states encountered during the demonstrations, significantly improving sample efficiency with much fewer demonstrations. Beyond reverse curriculum strategies, some approaches \cite{liu2021curriculum, yengera2021curriculum} propose ranking-based methods to iteratively select demonstrations from an offline pool for policy updates. However, most of these approaches rely on \textit{offline demonstrations} to construct curricula, assuming the distribution and contents of these demonstrations are known to the robot before the training starts. By contrast, our work focuses on automating curriculum iteration via \textit{online demonstrations}, the distribution of which is dynamic and can be iteratively queried by the robot during the learning process. Furthermore, we also examine how a curriculum learning based algorithm, which aligns with the intuitive learning patterns of humans, may affect human teaching—an area largely overlooked by previous research.

\subsection{Active Learning from Demonstrations}
To address the issue of covariance shift in conventional Learning from Demonstration (LfD) approaches, efforts have been made \cite{10309326, hou2023shaping} to allow the learning agent to take a more active role by querying demonstrations from human demonstrators. Some work \cite{silver2012active} optimizes the demonstration queries by framing them as an uncertainty reduction process in a supervised learning setting. More recent efforts \cite{chen2020active, chenactive} have introduced active LfD into sequential decision-making scenarios, where a Reinforcement Learning (RL) agent uses its Q-value function to construct various uncertainty measures to optimize the sequential querying of state-action pairs. However, querying isolated state-action pairs, similar to DAgger \cite{hoque2021lazydagger, kelly2019hg, ross2011reduction}, often leads to frequent context switches, which can increase the cognitive load on human demonstrators \cite{hou2024give}. To mitigate this, some approaches have focused on querying episodic demonstrations to reduce context switches \cite{hou2024give} and incorporating the human costs—such as time, effort, and failure recovery—into the query optimization process \cite{rigter2020framework}. Despite these advancements, most of these works overlook how human teaching can be influenced by the active query strategy of the learning agent. Although works like \cite{schrum2022mind, schrum2023reciprocal} actively provide robot feedback to human teachers to better shape their teaching strategy, they primarily focus on the sub-optimality of human feedback in terms of the timing aspect. By contrast, our work is focused on the distribution aspect of human demonstrations and explores how our active curriculum LfD algorithm can benefit both robot learning and human teaching. While \cite{sakr2023can} also aims to influence human teachers to select demonstrations that better support robot learning, their approach relies on visual feedback from AR devices and has only been tested on simple 2D navigation tasks. In contrast, our method requires no additional visual feedback support and is evaluated on more complex manipulation tasks involving object interactions.

\section{Methodology}
Our method aims to generate demonstration queries that benefit both robot learning and human teaching. Inspired by Curriculum Learning, which scaffolds tasks into progressively harder subtasks to support learning, we design an active LfD algorithm that leverages this intuitive progression to optimize the sequence of online demonstration queries.

\subsection{Preliminaries}
We formulate the problem of active LfD as a Markov Decision Process (MDP) represented by $(S, A, r, \gamma, P, \rho_0)$, where $S$ denotes the state space, $A$ is the action space, $r(s_t, a_t): S \times A \rightarrow \mathbb{R}$ is the reward function, $\gamma$ is the discount factor, $P(s_{t+1} | s_t, a_t)$ is the environment transition function, and $\rho_0$ is the initial state distribution. The goal is to optimize a RL policy $\pi$ that maximize the expected sum of discounted rewards $\mathbb{E}_{s_0 \sim \rho_0}[\sum_{t=0}^{T-1} \gamma^t r(s_t, \pi(s_t))]$ for episodes with a maximum length of $T$ steps.

Furthermore, we assume that the RL agent can also query a human demonstrator for a sequence of $N_d$ episodic online demonstrations during the policy training process. Each online demonstration $\xi^i=\{(s_t^i, a_t^i, r_t^i, s_{t+1}^i)\}_{t=0}^{T_i}$ consisting of $T_i$ steps will be iteratively added to the demo buffer $\mathcal{D}$ after it is queried by the RL agent. These off-policy demonstration data will be used together with the roll-out data generated by the RL agent itself to update its policy via the underlying LfD algorithm (e.g., Advantaged Weighted Actor-Critic \cite{nair2020awac}).

\subsection{Background on Advantage Weighted Actor-Critic}\label{AWAC}
In this work, we utilize Advantage Weighted Actor-Critic (AWAC) \cite{nair2020awac}, an advanced LfD algorithm, to update the policy with human demonstrations. More specifically, AWAC builds upon the Soft Actor-Critic (SAC) \cite{haarnoja2018soft} as its underlying off-policy RL algorithm and models the action-value function and policy via neural networks, denoted as $Q_{\phi}(s, a)$ and $\pi_{\theta}(s|a)$ respectively.

To update the action-value function $Q_{\phi}(s, a)$, AWAC optimizes its parameters $\phi$ via:
\begin{equation}\label{q}
    \phi_k = \arg\min_{\phi} \mathbb{E}_{D} \left[ (Q_{\phi}(s, a) - y)^2 \right],
\end{equation}
where
\begin{equation}
    y = r(s,a) + \gamma \mathbb{E}_{s', a'} \left[ Q_{\phi_{k-1}(s', a')} \right].
\end{equation}
And to update the policy network $\pi_{\theta}$, AWAC optimizes its parameters $\theta$ by:
\begin{equation}\label{pi}
    \theta_{k+1} = \arg\max_{\theta} \mathbb{E}_{s,a \sim \beta} \left[\log\pi_{\theta}(s|a) \exp \left( \frac{1}{\lambda} A^{\pi_{\theta}}(s,a) \right) \right],
\end{equation}
where $\beta$ is the replay buffer defined as $\beta = \mathcal{D} \cup \mathcal{R}$, including the data from both the demo buffer $\mathcal{D}$ and self roll-out buffer $\mathcal{R}$. $\lambda$ is a hyperparameter and $A^{\pi_k}$ is the advantage under the policy $\pi_{\theta}$ defined as $A^{\pi_{\theta}}(s,a) = Q^{\pi_{\theta}}(s,a) - V^{\pi_{\theta}}(s)$, which functions as a dynamic weight to encourage the policy update towards actions estimated to be of higher action-values.

\subsection{Active Curriculum Learning from Online Demonstrations}
Inspired by \cite{taoreverse}, we construct a sequence of curricula with gradually increasing difficulties by resetting the environment to states that are progressively farther away from the task goal area and closer to the initial state space of the original task. Each curriculum $c_i$ itself is essentially an initial state distribution around a center state $s^{c_i}$, denoted as $\rho_0(s_0 | s^{c_i})$. For instance, if the center state $s^{c_i}$ is defined as the coordinate $(x, y)$ of a cube for a robot arm to pick up, the corresponding curriculum $c_i$ can be a uniform distribution $\rho_0(s_0 | s^{c_i})$ of the cube positions that are within a circle area centered at $s^{c_i}$. Based on the contexts, we may use $c_i$ and $\rho_0(s_0 | s^{c_i})$ interchangeably. 

\textbf{Collect a base demo}. Before training starts, we first query an episodic demonstration whose initial state is randomly sampled from the initial state distribution $\rho_0$ specified by the original task. We refer to this demonstration as \textit{base demo} denoted as $\xi^{base}$ with a length of $T_{b}$ steps. This base demo provides a viable path to solve the task from the original initial state distribution $\rho_0$ and functions as the backbone to construct a sequence of curricula with increasing difficulties. With the base demo $\xi^{base}$, we initialize the base-demo stage $g \in \mathbb{N}$ with the value of $1$,  create the very first curriculum $c_0$ whose center state $s^{c_0} = s^{base}_{T_b - g}$ chosen as the last state of the base demo $\xi^{base}$, and add it to an empty candidate curricula list denoted as $\mathcal{C}$.

\textbf{Evaluate curricula difficulties}. For each iteration during the training process, we first evaluate the difficulty of each curriculum inside the candidate curricula list $\mathcal{C}$. For each curriculum $c_i$, we roll out the current policy $\pi_{\theta}$ for $N_{eval}$ episodes whose initial states $s_0 \sim \rho_0(s_0 | s^{c_i})$ and calculate the average success rate $q_{c_i}$. Similar to \cite{taoreverse}, we assign a \textit{reachability score} of $2$ if $q_{c_i} = 0$, a score of $3$ if $0 < q_{c_i} < w$ where $w$ is a given threshold of success rate, and a score of $1$ if $q_{c_i} \geq w$, aiming to encourage choosing the curriculum that is at the edge of the capacity of the current policy.

\textbf{Select the current curriculum}. After evaluating the reachability scores for each candidate curriculum, we select the curriculum $c^*$ used to update the policy as the one of the highest reachability score. If there are multiple curricula of the same highest score, we randomly sample one from them as the current curriculum $c^*$.

\textbf{Update the policy and selected curriculum}. To update the policy $\pi_{\theta}$, we roll out the current $\pi_{\theta}$ for $N_{train}$ episodes whose initial states $s_0 \sim \rho_0(s_0 | s^{c^*})$ and update $Q_{\phi}$ and $\pi_{\theta}$ via \eqref{q} and \eqref{pi} respectively after each episodic roll-out, with the number of updates matching the number of steps in the roll-out. And to update the selected curriculum $c^*$ after all policy updates, we do a post-update curriculum evaluation by measuring the average success rate $q_{c^*}'$ under the updated policy $\pi_{\theta}$ for $N_{eval}$ episodes with $s_0 \sim \rho_0(s_0 | s^{c^*})$. We will remove $c^*$ from the candidate curricula list $\mathcal{C}$ if $q_{c^*}' \geq w$. Furthermore, if the center state $s^{c^*}$ of $c^*$ is from the base demo and base-demo stage $g$ has not reached the start of the base demo (i.e., $g < T_b$), we will update the base-demo stage $g$ to $\min(g + \delta_g, T_b)$, with the stage update size $\delta_g \in \mathbb{N}$ as a pre-defined hyperparameter. We then create a new curriculum $c_i$ with its center state as $s^{c_i} = s^{base}_{T_b - g}$ and add it to $\mathcal{C}$.

\textbf{Expand curricula with online demonstrations}. Till now, all curricula within the candidate curricula list $\mathcal{C}$ are generated with their center states obtained from the base demo. This may limit the policy exploration only to the state areas that are adjacent to the single base demo $\xi^{base}$. To encourage a broader state-space coverage while maintaining a from-easy-to-hard exploration process, we maintain a separate query-demo stage $\widetilde{g}$ with the same stage update size $\delta_g$ to decide what online demonstration to query from the human demonstrator at an interval of $N_q$ ($N_q = n N_{train}, n \in \mathbb{N}$) episodes of policy updates. At each query-demo stage $\widetilde{g}$, the agent actively queries an episodic demonstration $\xi^{online}$ with its initial state $\widetilde{s_0} \sim \widetilde{\rho}_0(s_0 | s^{base}_{T_b - \widetilde{g}})$. Different from $\rho_0(s_0 | s^{c_i})$, $\widetilde{\rho}_0(s_0 | s^{base}_{T_b - \widetilde{g}})$ represents a distribution that will sample states of the same difficulty level to start from as the $s^{base}_{T_b - \widetilde{g}}$. In the example of cube-pick-up task, $\widetilde{\rho}_0(s_0 | s^{base}_{T_b - \widetilde{g}})$ can be a uniform distribution for all 2D cube positions that are of the same euclidean distance to the goal area as that between the demo cube position $s^{base}_{T_b - \widetilde{g}}$ and the goal area, i.e., an arc centered at the goal area with a radius of the euclidean distance between $s^{base}_{T_b - \widetilde{g}}$ and the goal. After the online demo is collected, we create a new curriculum $c_i$ with $\widetilde{s_0}$ as the center state and add it to $\mathcal{C}$. We summarize the pseudo-code as in Algorithm \ref{alg}. 

\begin{algorithm}[t!]
\caption{Active Curriculum Learning from Online Demonstrations}\label{alg}
\begin{algorithmic}[1]
\Require evaluation interval $N_{eval}$, update interval $N_{train}$, score threshold $w$, stage update size $\delta_g$, demo budget $N_d$

\State Randomly initialize actor-critic networks $Q_{\phi}$ and $\pi_{\theta}$
\State Initialize demo buffer $\mathcal{D}$ and self roll-out buffer $\mathcal{R}$
\State Collect a base demo $\xi^{base}$ of length $T_b$
\State Update demo buffer $\mathcal{D} \gets \mathcal{D} \cup \{\xi^{base}\}$
\State Initialize base-demo stage $g$ and query-demo stage $\widetilde{g}$
\State Initialize curriculum $c_0$ with center state $s^{c_0} \gets s^{base}_{T_b - g}$
\State Initialize candidate curricula list $\mathcal{C} \gets \{c_0\}$
\State Queried demo number $n_d \gets 0$

\For{iteration $k \in \{1, 2, ...\}$}
    \If{$\mathcal{C}$ is not empty}
        \For{candidate curriculum $c_i \in \mathcal{C}$}
            \State rollout $\pi_{\theta}$ for $N_{eval}$ episodes, $s_0 \sim \rho_0(s_0 | s^{c_i})$
            \State update self roll-out buffer $\mathcal{R}$
            \State calculate average success rate $q_{c_i}$
            \State convert $q_{c_i}$ to the reachability score $q_{c_i}^{score}$
        \EndFor
    
        \State select curriculum $c^* \gets \argmax_{c_i \in \mathcal{C}}{q_{c_i}^{score}}$
        \State rollout $\pi_{\theta}$ for $N_{train}$ episodes, $s_0 \sim \rho_0(s_0 | s^{c^*})$
        \State update self roll-out buffer $\mathcal{R}$ 
        \State update $Q_{\phi}$ and $\pi_{\theta}$ via \eqref{q} and \eqref{pi}
        \State calculate post-update success rate $q_{c^*}'$ for $c^*$
        \If{$q_{c^*}' \geq w$}
            \State remove $c^*$ from $\mathcal{C}$
        \EndIf
        \If{$c^*$ is from base demo and $g < T_b$}
            \State update base-demo stage $g \gets \min(g + \delta_g, T_b)$
            \State create a new curriculum $c_i$ with $s^{c_i} \gets s^{base}_{T_b - g}$
            \State $\mathcal{C} \gets \mathcal{C} \cup \{ c_i \}$
        \EndIf
        \If{$k \, \% \, N_{train} == 0 $ and not reach budget $N_d$}
            \State query a demo $\xi^{online}$ with $\widetilde{s_0} \sim \widetilde{\rho}_0(s_0 | s^{base}_{T_b - \widetilde{g}})$
            \State update demo buffer $\mathcal{D}$
            \State create a new curriculum $c_i$ with $s^{c_i} \gets \widetilde{s_0}$
            \State update $\mathcal{C} \gets \mathcal{C} \cup \{ c_i \}$ and $\widetilde{g} \gets \min(\widetilde{g} + \delta_g, T_b)$
        \EndIf
    \Else
        \State rollout $\pi_{\theta}$ with $s_0 \sim \rho_0$
        \State update $\mathcal{R}$, $Q_{\phi}$ and $\pi_{\theta}$
    \EndIf
\EndFor
\end{algorithmic}
\label{algorithm1}
\end{algorithm}

\section{Experimental Setup}
We design experiments to evaluate our method's learning performance and its impact on human teaching. For learning performance, we test four increasingly difficult robotic tasks in simulation, comparing success rates and sample efficiency with other LfD baselines. To assess human teaching, we conduct a user study comparing our method with another active LfD approach in terms of teaching performance, adaptivity, and transferability.

\subsection{Simulation Experiments of Robot Learning}
\subsubsection{Task Environments}

\begin{figure}[t!]
  \centering
  \includegraphics[width=0.8\linewidth]{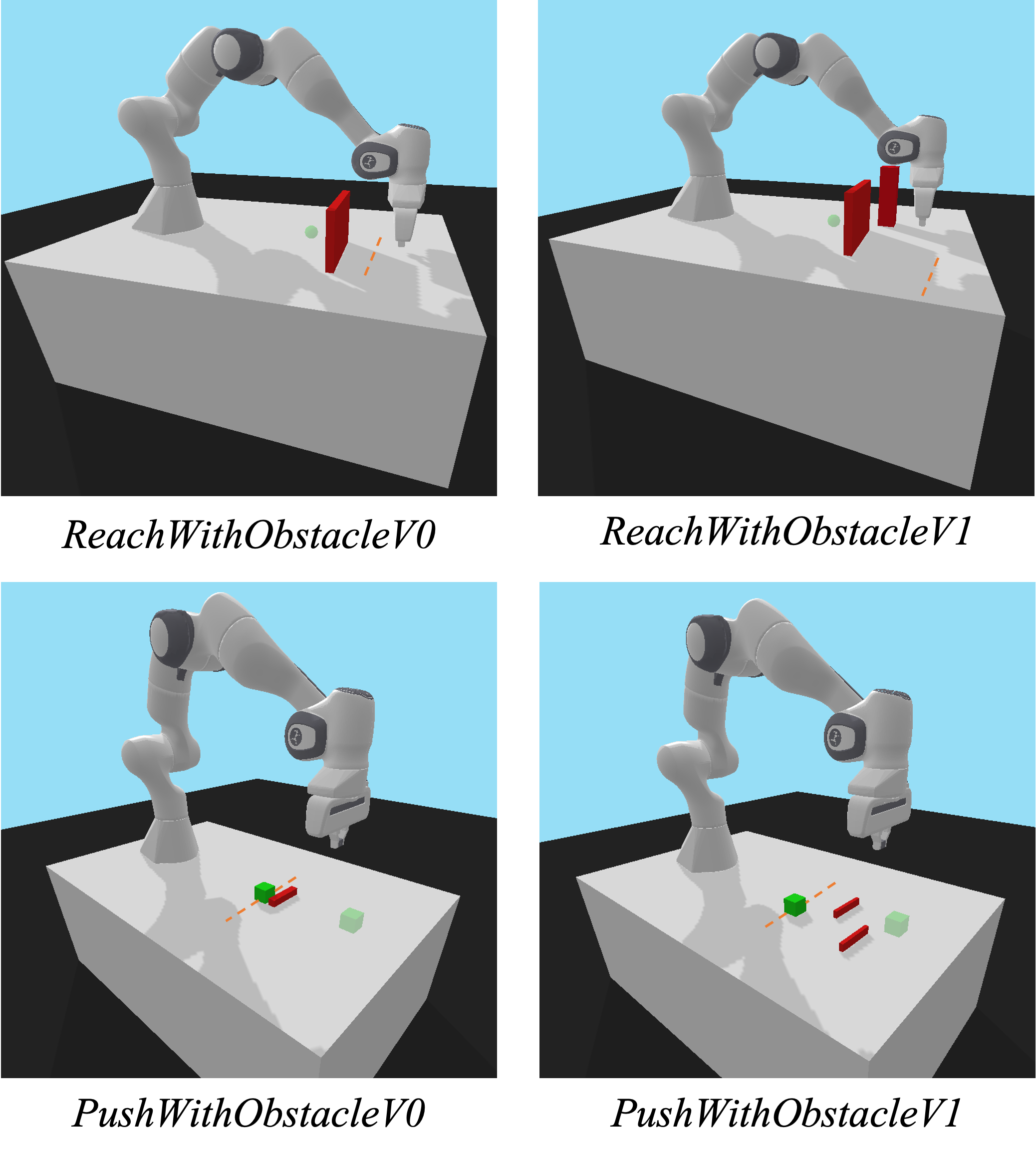}
  \caption{Simulation task environments, with orange lines indicating starting positions for the cube or end effector.}
  \label{sim envs}
  \vspace{-3mm}
\end{figure}

\begin{figure*}[t!]
  \centering
  \includegraphics[width=1.0\linewidth]{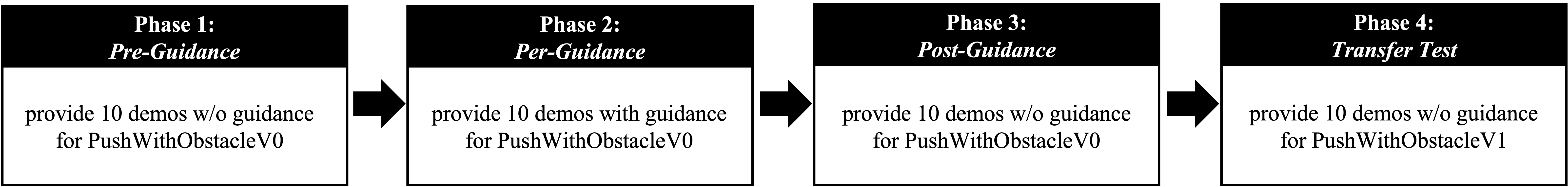}
  \vspace{-6mm}
  \caption{Procedures of the user study of human teaching}
  \label{user study procedures}
  \vspace{-3mm}
\end{figure*}

Based on the panda-gym environment \cite{gallouedec2021pandagym}, we design $4$ robotic tasks with increasing difficulties as shown in Fig. \ref{sim envs}, namely \textit{ReachWithObstacleV0}, \textit{ReachWithObstacleV1}, \textit{PushWithObstacleV0}, and \textit{PushWithObstacleV1}. For the task of ReachWithObstacleV0, the robot end effector moves from a random position sampled from a predefined starting line of a length of $0.3$m and aims to reach the fixed goal position right behind a obstacle wall. The robot arm is expected to reach the goal with as few steps as possible while not crushing into the obstacle. It will receive a $+1000$ reward if arriving at the goal, a $-1000$ reward if it touches the wall, or a $-1$ reward otherwise. An episode will terminate immediately if the end effector crushes into the wall, arrives at the goal, or it reaches the maximum episode length of $120$ environment steps. For the task of ReachWithObstacleV1, it is similar to ReachWithObstacleV0 except that there is another column obstacle that the robot needs to avoid touching. For the task of PushWithObstacleV0, the robot arm starts from a pre-defined neutral pose and aims to push a $0.04$m-sized cube across the table surface to arrive at a fixed goal position without the cube touching the obstacle. The initial position of the cube is randomly sampled from a predefined starting line of a length of $0.3$m. The definition of the reward function and the terminal condition are the same as ReachWithObstacleV0. And for the task of PushWithObstacleV1, it is similar to PushWithObstacleV0but, but with the addition of an extra obstacle that the cube needs to avoid.

\subsubsection{Baselines}
To evaluate learning performance, we compare our method with $3$ LfD baselines in terms of final success rate of the converged policy and sample efficiency evaluated over $10$ random seeds. We consider a policy as converged when the variance in its success rate remains within $5\%$ over $10 \times 10^3$ consecutive steps. Specifically, we compare with $2$ popular LfD approaches using offline demonstrations and $1$ active LfD method using online demonstrations:
\begin{itemize}
    \item \textbf{DDPGfD-BC}: An LfD method using offline demonstrations that integrates a Behavioral Cloning term into the policy update to further encourage imitation \cite{nair2018overcoming}. We use the same training hyperparameter values as specified in the original work.
    \item \textbf{AWAC}: Another canonical LfD method utilizing offline demonstrations as mentioned in Section \ref{AWAC}. We use the same network architectures for the policy and action-value functions as in the original work, along with the default hyperparameter values specified in the paper.
    \item \textbf{EARLY}: A latest active LfD method leveraging episodic online demonstrations queried based on a TD-error-driven uncertainty measure \cite{hou2024give}. All the training hyperparameters are set as the same values as in the original work.
\end{itemize}

For our method, we set the evaluation interval $N_{eval}=10$, update interval $N_{train}=20$, score threshold $w=0.7$, stage update size $\delta_g = 10$, and demo budget $N_d=10$. We employ the same actor-critic structure as the AWAC and utilize layer normalization along with balanced sampling from the self-rollout buffer and the demonstration buffer as in \cite{ball2023efficient} to help stabilize policy updates.

\subsubsection{Human Demonstration Collection}
For each task, we use a joystick to collect a pool of 60 successful episodic demonstrations from the same human expert. The initial state of each demonstration is randomly sampled from the initial state distribution. For both DDPGfD-BC and AWAC, we randomly select $N_d=10$ demonstrations from the pool and load them into the replay buffer before training for the corresponding task. For EARLY, when an online query is generated, we select the demonstration from the pool whose initial state is closest to the queried state and use that demonstration as the queried online demonstration. For our method, when an online query is generated, we iterate through each demonstration to find the state closest to the queried one and use the partial demonstration starting from that state until the end as the queried online demonstration.

It should be noted that these demonstration pools serve as a methodological trick, acting as a proxy for an unknown human-expert policy. They provide human demonstrations and pass them to the robot upon request. The mechanism by which each demonstration is selected from the pool remains unknown to the robot, simulating the presence of a real human demonstrator throughout the training process. In this setup, it is equivalent to a human demonstrator being continuously available, iteratively offering joystick-controlled demonstrations whenever the robot makes a query.

\subsection{User Study of Human Teaching}
\subsubsection{Participants}
To investigate human teaching under active LfD algorithms, we conducted a between-subject user study with $26$ human participants ($14$ male, $10$ female, and $2$ other; $15$ aged between $18-29$ and $11$ aged between $30-39$; $8$ of no experience of machine learning, $8$ of some experience of machine learning and $10$ of extensive experience). We randomly divided the participants into two groups: one group for experiments using our method as the underlying active LfD algorithm, and the other for experiments using the EARLY method. We recruited these participants via poster advertisement on campus under the approval of our faculty research ethics board. We obtained informed consent for the experiment and data collection from each participant prior to the start of the study. As compensation, each participant received a €10 gift card for attending the experiments.

\subsubsection{Procedures}
To investigate how human teaching evolves under the guidance of active LfD algorithms, we designed a user study consisting of $4$ phases, as shown in Fig. \ref{user study procedures}. We did not include LfD methods with offline demonstrations (i.e., DDPGfD-BC and AWAC) in the user study because human teaching (i.e., demonstration collection) concludes before robot learning in these cases and there is no influence from robot guidance involved in this process. 

Before all experiments start, participants undergo a 5-minute training session to familiarize themselves with the joystick interface. They may pass it if they successfully provide demonstrations from randomly assigned initial positions five times consecutively, or if it reaches the $5$-minute time limit.

\textbf{Phase 1: Pre-Guidance}. 
In this phase, participants teach the robot arm to perform the task of \textit{PushWithObstacleV0} in simulation. More specifically, participants provide $10$ successful demonstrations that they believe to be most beneficial for robot learning, \textit{without any guidance from the robot}. For each demonstration, participants first use a joystick to select the starting positions of the cube and the robot gripper, then provide the corresponding demonstration after confirming their selections. After all demonstrations are provided, participants complete an open-ended questionnaire to describe the strategies they used in Phase 1 to select the initial positions.


\textbf{Phase 2: Per-Guidance}. In this phase, the robot leverages an active LfD algorithm (either ours or EARLY) to generate online queries during the training process and iteratively asks for $10$ successful online demonstrations from each participant for the task of \textit{PushWithObstacleV0}. For each query, the environment resets the cube and the gripper to the queried positions and participants then use the joystick to provide the corresponding demonstration. After each demonstration, we visualize the queried starting position in the environment to help participants keep track of the robot's queries and better reflect on its query strategy. After all $10$ demonstrations are collected, participants complete a standard NASA-TLX questionnaire to quantitatively assess their perceived workload under the algorithm. They also fill out an open-ended questionnaire to describe the robot query strategy they perceive for selecting the initial positions in Phase 2.

\textbf{Phase 3: Post-Guidance}. In this phase, human participants again teach the robot how to perform the task of \textit{PushWithObstacleV0} by providing $10$ successful demonstrations to the robot via the joystick. Similar to Phase 1, the participants select the initial positions of both the cube and the gripper without the guidance of the robot. After all $10$ demonstrations are provided, participants fill out an open-ended questionnaire to describe the strategy they employed during Phase 3.

\textbf{Phase 4: Transfer Test}. In this phase, human participants teach the robot to perform the \textit{PushWithObstacleV1} task, which they have not previously taught with the robot's guidance. Similar to Phase 3, the participants use the joystick to select the starting positions of the cube and gripper without the robot guidance, and provide $10$ successful demonstrations. 

To train policies with demonstrations from Phases 1, 3, and 4, we load the human-selected demonstrations and sequentially feed them to the active LfD algorithm (EARLY or ours) during queries. For Phase 2, we save the critic-actor model checkpoints after participants provide all $10$ guided demonstrations, then resume training from the checkpoint post-experiments.

\subsubsection{Metrics on Human Teaching}
\begin{itemize}
    \item \textbf{Human teaching performance}: To evaluate human teaching performance, we measure the total human time spent by each participant in providing demonstrations during Phase 2 and the number of failed demonstration attempts during their demonstrations in this phase. Total human time is measured from the start of training until all $10$ successful demonstrations are collected. Additionally, we also measure the perceived workload assessed by the standard NASA-TLX questionnaire. 
    
    \item \textbf{Human teaching adaptivity}: We evaluate human teaching adaptability by measuring the increase in the final success rate of the converged policy in Phase 3 compared to Phase 1, i.e., the \textit{success rate increment} from the pre-guidance phase to the post-guidance phase. Additionally, we report the final success rate of the converged policy in Phase 3, referring to it as \textit{post-guidance teaching efficacy}.

    \item \textbf{Human teaching transferability}: We evaluate human teaching transferability by measuring the final success rate of the converged policy trained with demonstrations selected by human teachers for a task they have not previously taught under robot guidance but have taught a similar task with robot guidance. 
    
\end{itemize}

\section{Results and Discussion}
\subsection{Results for Simulation Experiments}

\begin{figure*}[t!]
  \centering
  \includegraphics[width=1.0\linewidth]{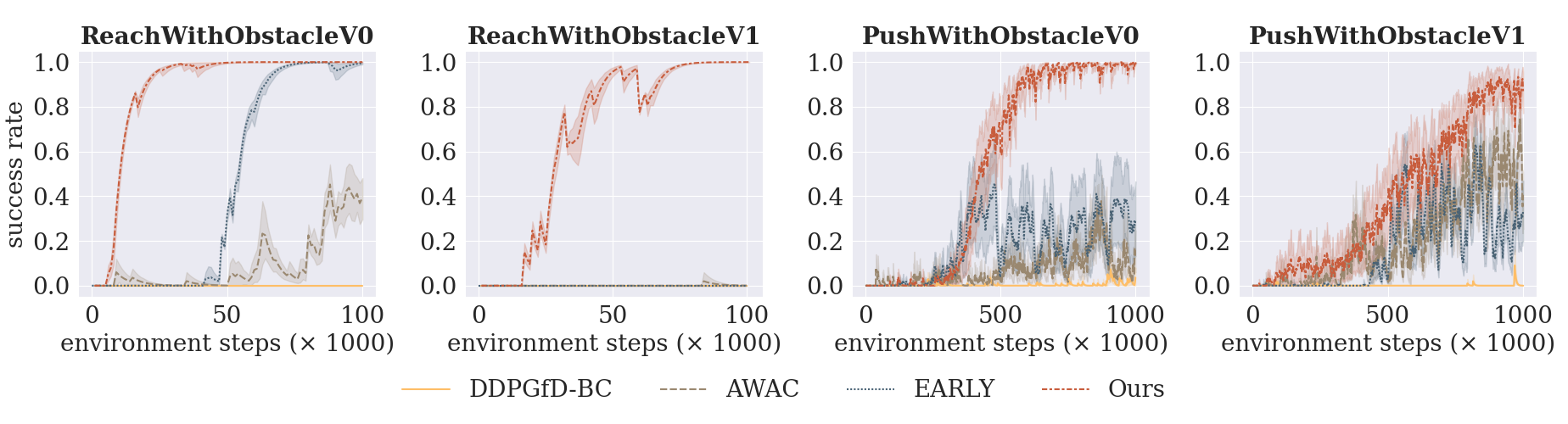}
  \vspace{-5mm}
  \caption{Results of simulation experiments for four robotic tasks.}
  \label{sim res}
  \vspace{-4mm}
\end{figure*}

As shown in Fig. \ref{sim res}, our method outperforms all other baselines in all $4$ tasks. More specifically, for ReachWithObstacleV0, both our method and EARLY converge to $100\%$ success rate. However, our method only consumes around $22 \times 10^3$ environment steps for convergence, $68.6\%$ faster than EARLY that takes about $70 \times 10^3$ steps to converge. By contrast, AWAC only achieves $40\%$ success rate and DDPGfD-BC fails to solve the task. For  ReachWithObstacleV1, only our method manages to converge to $100\%$ success rate while all the baselines fail to solve the task. For PushWithObstaclev0, our method also manages to converge to $100\%$ success rate. By contrast, EARLY only achieves a success rate around $40\%$, AWAC converges around $30\%$, and DDPGfD-BC achieves around $5\%$ success rate, all worse than our method. For PushWithObstacleV1, our method converges to over $90\%$ success rate. By comparison, AWAC only converges around $70\%$ success rate, EARLY achieves around $40\%$ success rate, DDPGfD-BC fails to solve the task. These results indicate that our method significantly improves the robot learning performance in both final policy performance and sample efficiency compared with other LfD baselines.

\subsection{Results for User Study Experiment}

\begin{figure}[t!]
  \centering
  \includegraphics[width=0.9\linewidth]{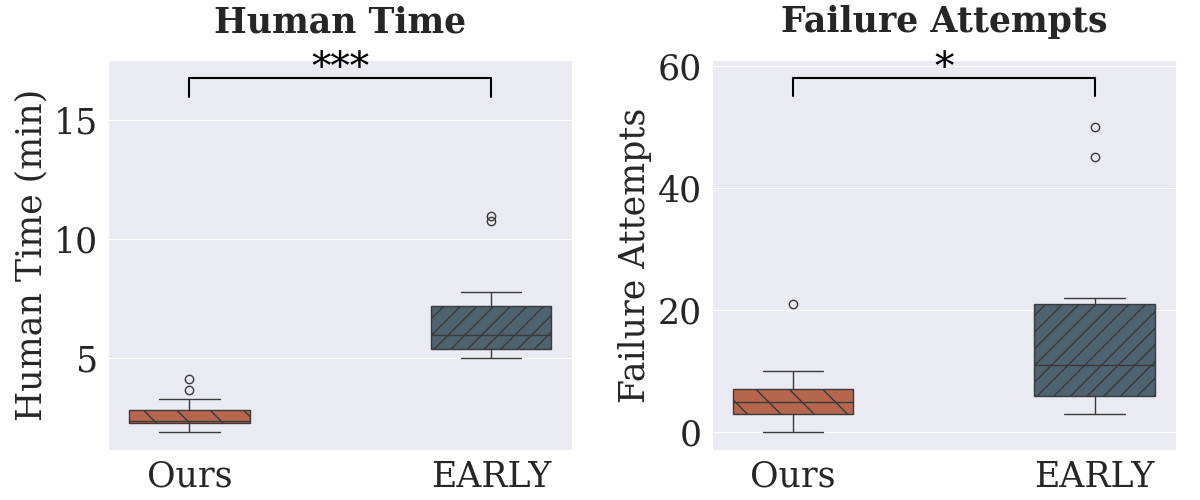}
  \caption{Results of human teaching performance in Phase 2.}
  \label{time and failures}
  \vspace{-3mm}
\end{figure}

\begin{figure}[t!]
  \centering
  \includegraphics[width=1.0\linewidth]{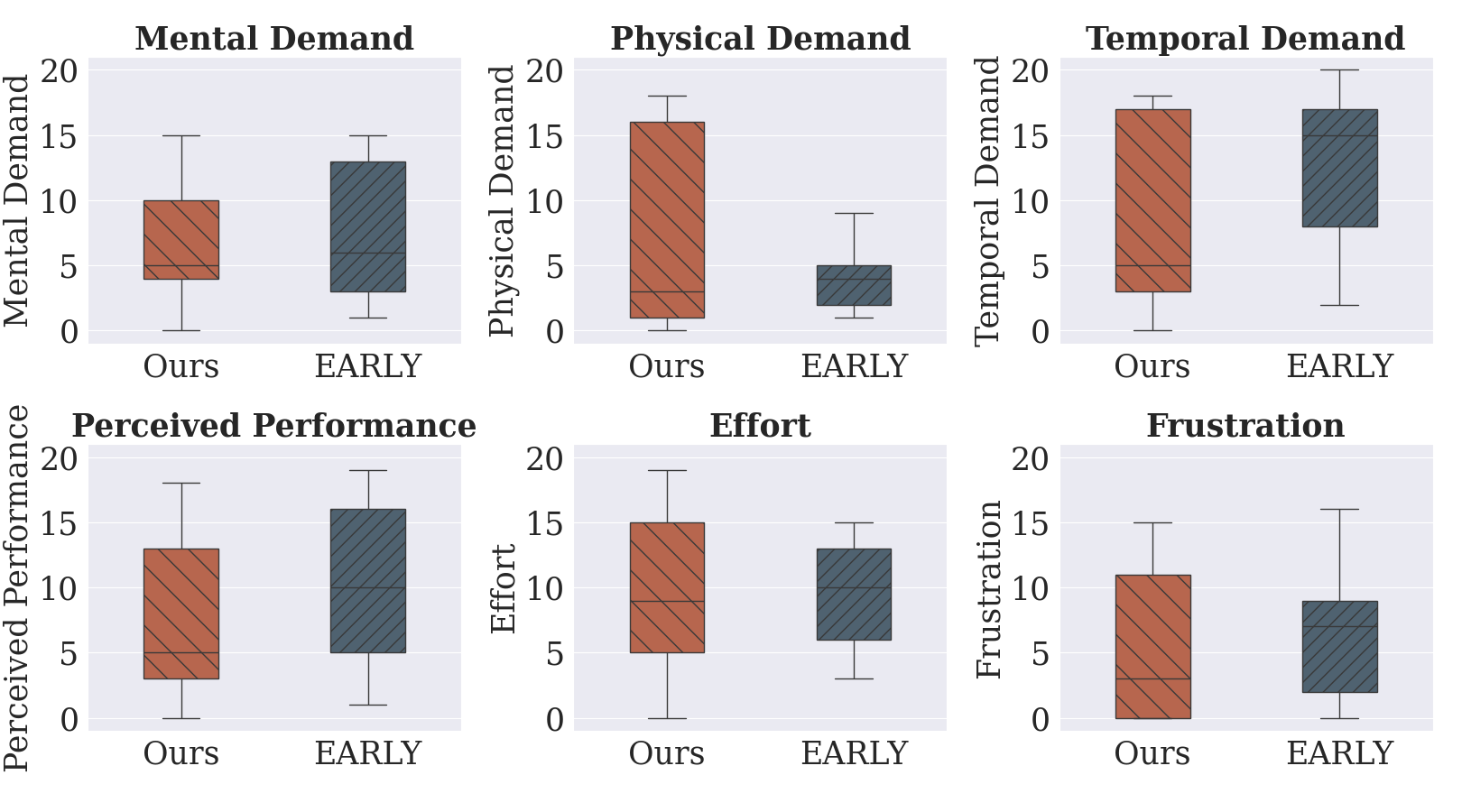}
  \vspace{-6mm}
  \caption{Results of perceived workload in Phase 2.}
  \label{workload}
  \vspace{-3mm}
\end{figure}

\subsubsection{Human Teaching Performance}
Since the total human time and number of failed demonstration attempts do not pass the Shapiro-Wilk test for normality, we conduct Mann-Whitney U tests to compare both metrics in Phase 2 between our algorithm and the EARLY condition. As in Fig. \ref{time and failures}, the results show that our method (median=$2.35$, IQR=$[2.23, 2.78]$) takes significantly less human time than EARLY (median=$5.95$, IQR=$[5.36, 7.18]$), $U =0.0$, $p < .001$ (two-sided). Furthermore, our method (median=$5.0$, IQR=$[3.0, 7.0]$) takes significantly fewer failed demonstration attempts compared with EARLY (median=$11.0$, IQR=$[6.0, 21.0]$), $U =41.0$, $p < .05$ (two-sided). These results indicate that human teaching performance is more time-efficient and less prone to demonstration failures when guided by our algorithm compared with the EARLY baseline.


Since most perceived workload metrics, except for effort, in Phase 2 do not pass the Shapiro-Wilk test for normality, we use Mann-Whitney U tests to compare them between our algorithm and EARLY. For effort, which passes the normality test, we conduct an independent samples t-test. The results (Fig. \ref{workload}) show that there is no significant difference between our method and EARLY in terms of mental demand (median=$5.0$, IQR=$[4.0, 10.0]$ for ours and median=$6.0$, IQR=$[3.0, 13.0]$ for EARLY, $U =84.5 $, $p=1.0$), physical demand (median=$3.0$, IQR=$[1.0, 16.0]$ for ours and median=$4.0$, IQR=$[2.0, 5.0]$ for EARLY, $U=88.0 $, $p=0.88$), temporal demand (median=$5.0$, IQR=$[3.0, 17.0]$ for ours and median=$15.0$, IQR=$[8.0, 17.0]$ for EARLY, $U =59.0$, $p=0.20$), perceived performance (median=$5.0$, IQR=$[3.0, 13.0]$ for ours and median=$10.0$, IQR=$[5.0, 16.0]$ for EARLY, $U =55.0 $, $p=0.14$), and frustration (median=$3.0$, IQR=$[0.0, 11.0]$ for ours and median=$7.0$, IQR=$[2.0, 9.0]$ for EARLY, $U =69.5 $, $p=0.45$). And there is no significant difference in effort (M=$9.70$, SD=$6.69$ for ours and M=$9.77$, SD=$4.21$ for EARLY, $t(24)=-0.04, p=0.97$). These results indicate that our method shares the same level of perceived workload as EARLY.

Combining the results of perceived workload and failed demonstration attempts, we find that human participants have to undergo on average at least $5$ failures in total to provide $10$ successful demonstrations. Due to this significant failure rate, we hypothesize that the perceived workload reflects more on the limited usability of the data collection interface (i.e., joystick) rather than the underlying algorithm itself. This suggests that interface usability may be a confounding factor that has overshadowed the effects of the underlying algorithm. Indeed, from the feedback we received from the human participants, many of them mentioned that their user experience was largely influenced by the unfriendly usability of the joystick interface. We leave it as future work to further investigate the impact of interface usability on human teaching and optimize the interface design to facilitate the teaching process. Overall, our method is more time-efficient and less prone to demonstration failures while maintaining the same perceived workload as the EARLY baseline, indicating better teaching performance.

\begin{figure}[t!]
  \centering
  \includegraphics[width=1.0\linewidth]{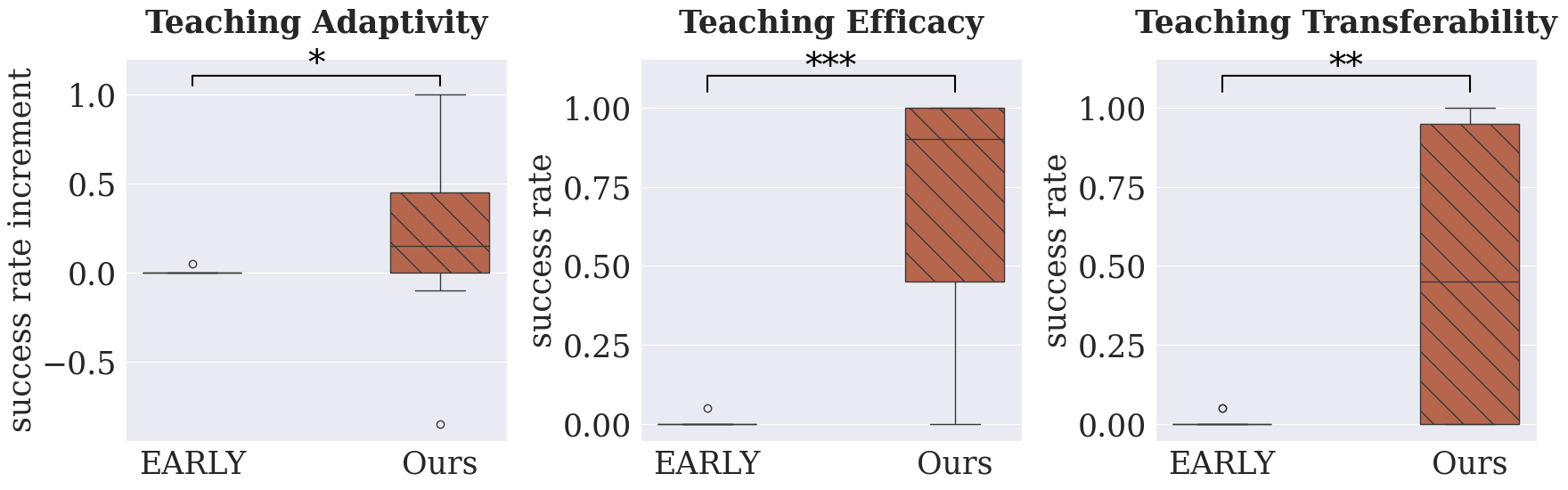}
  \vspace{-4mm}
  \caption{Results of teaching adaptivity, post-guidance teaching efficacy, and teaching transferability.}
  \label{adap and transfer}
  \vspace{-3mm}
\end{figure}

\subsubsection{Human Teaching Adaptivity}

Since the data fail to pass the Shapiro-Wilk test for normality, we use a Mann-Whitney U test to compare the teaching adaptability of human participants between our method and EARLY. The results (Fig. \ref{adap and transfer}) show that there is a significant difference between our method (median=$0.15$, IQR=$[0.0, 0.45]$) and EARLY (median=$0.0$, IQR=$[0.0, 0.0]$), $U =128.0$, $p<.05$. Furthermore, since the data for post-guidance teaching efficacy also failed the Shapiro-Wilk test for normality, we perform a Mann-Whitney U test to compare it between our method and EARLY. As in Fig. \ref{adap and transfer}, there is a significant difference in post-guidance teaching efficacy between our method (median=$0.9$, IQR=$[0.45, 1.0]$) and EARLY (median=$0.0$, IQR=$[0.0, 0.0]$), $U =162.0 $, $p<.001$. These results indicate that human participants can significantly better adapt their teaching strategies to the robot learning after being guided by our method compared with the EARLY baseline, leading to both significantly improved robot learning and greater increments in learning performance.



\begin{figure}[t!]
\centering
\begin{subfigure}{0.4\linewidth}
  \centering
  \includegraphics[width=0.9\textwidth]{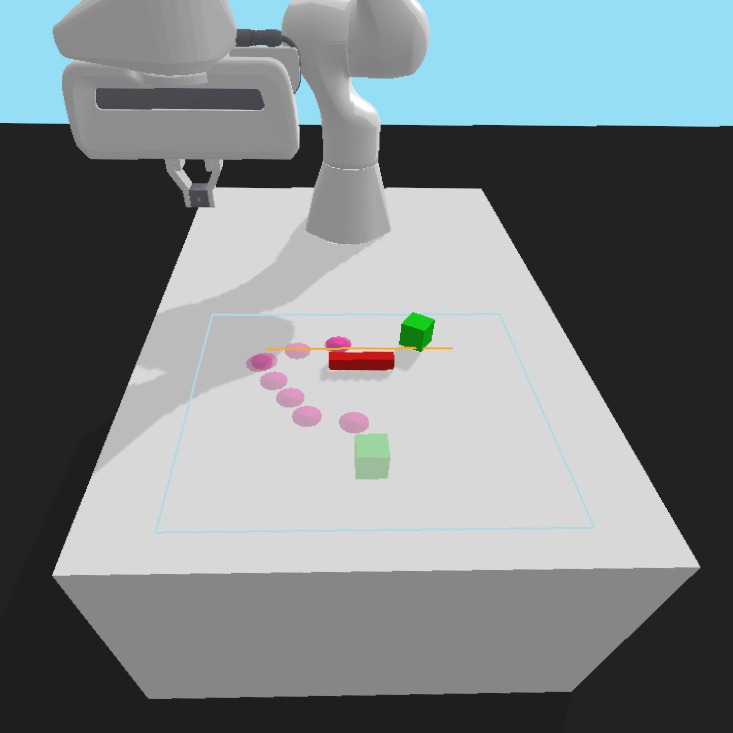}
  \caption{Query trace of ours}
\end{subfigure}
\begin{subfigure}{0.4\linewidth}
  \centering
  \includegraphics[width=0.9\textwidth]{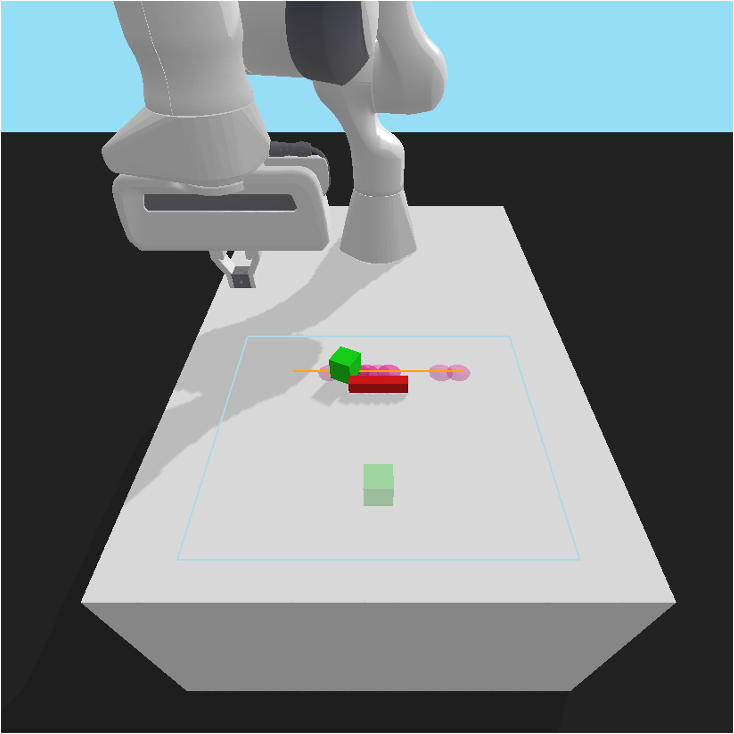}
  \caption{Query trace of EARLY}
\end{subfigure}
\caption{Example traces of robot queries in Phase 2, with pink circles marking the queried initial cube positions.}
\label{trace}
\vspace{-3mm}
\end{figure}

To examine how human teaching strategies evolve from the pre-guidance phase (Phase 1) to the post-guidance phase (Phase 3), we qualitatively analyze responses to open-ended questions about teaching strategies collected in Phases 1, 2, and 3. The results show that $61.5\%$ of participants guided by our algorithm in Phase 2 noted that the underlying query algorithm seemed to “learn the task incrementally” and “explore areas progressively further from the goal,” aligning with the principle of learning from easy to hard. Additionally, $53.8\%$ of participants mentioned that they adapted their strategies in Phase 3, choosing situations with increasing difficulty. Indeed, as shown in Fig. \ref{trace}, the trace of robot queries generated by our method exhibits a clear from-easy-to-hard trend. In contrast, only $7.7\%$ of participants guided by EARLY in Phase 2 conceived the algorithm queried areas where it lacked confidence, and only $38.5\%$ reported adapting their teaching strategies in Phase 3, often without understanding why the robot needed assistance in these areas. These results suggest that participants find our algorithm's query strategy more intuitive and are qualitatively more proactive in adjusting their teaching strategies to better support the robot's learning after the guidance of our method.


\subsubsection{Human Teaching Transferability}
Since the data do not pass the Shapiro-Wilk test for normality, we conduct a Mann-Whitney U test to compare the post-guidance teaching transferability of human participants in Phase 4 between our method and EARLY. As shown in Fig. \ref{adap and transfer}, the results show that there is a significant difference between our method (median=$0.45$, IQR=$[0.0, 0.95]]$) and EARLY (median=$0.0$, IQR=$[0.0, 0.0]$), $U =139.0  $, $p<.01$. This indicates that with our algorithm human participants can better transfer their teaching strategy to the unseen task compared with EARLY.

\subsection{Pathway to Deployment}
In this work, we assume the availability of an environment reset function, allowing the environment to be reset to any state for demonstration queries. While this assumption may not be practical for real-world deployment of our algorithm, it can be mitigated with human assistance or predefined motion planners. Moreover, this work influences human teaching in an implicit way by embedding intuitive scaffolding concepts into the query generation process, implicitly prompting humans to better understand the robot's query strategy and adjust their teaching accordingly. This may get influenced by the usability of the data collection interface (e.g., joystick control) if it requires much effort for human teachers to become proficient while comprehending the robot guidance at the same time. Therefore, to deploy our algorithm in real-world scenarios, a more user-friendly and intuitive data collection interface may also be of great help to facilitate human teaching and decrease the distraction caused by the interface itself.

\section{Conclusions}
We present an active LfD algorithm that uses Curriculum Learning to optimize online demonstration queries, guiding teachers through progressively challenging scenarios. Evaluated on four sparse-reward robotic tasks, our method outperforms three LfD baselines in policy success rate and sample efficiency. A further user study shows it requires less human time, fewer failed demonstration attempts, and improves teaching adaptivity, post-guidance efficacy, and transferability compared to the active LfD baseline. For future work, we plan to explore the impact of demonstration interfaces on teaching and develop more informative metrics for assessing human factors in human teaching, particularly in scenarios involving active learning from human demonstrators.


\section{Acknowledgments}
Thanks to Konstantin Mihhailov for his assistance in conducting the human teaching user study.

\bibliographystyle{IEEEtran}
\balance
\bibliography{refs}

\end{document}